\documentclass[runningheads]{llncs}
\usepackage{graphicx}

\usepackage{tikz}
\usepackage{comment}
\usepackage{amsmath,amssymb} 
\usepackage{color}
\usepackage{booktabs}
\usepackage{wrapfig}
\usepackage{subfigure}
\usepackage{multirow}
\usepackage{algorithm}
\usepackage{algorithmic}
\usepackage[T1]{fontenc}
\usepackage{makecell}

\begin{document}
\pagestyle{headings}
\mainmatter

\title{Alleviating Robust Overfitting of Adversarial Training With Consistency Regularization} 


\titlerunning{Abbreviated paper title}
%
\author{Shudong Zhang\inst{1} \and
Haichang Gao\inst{1} \and
Tianwei Zhang\inst{2} \and
\\Yunyi Zhou\inst{1} \and
Zihui Wu \inst{1}
}
\authorrunning{S. Zhang et al.}
%
\institute{Xidian University \and
Nanyang Technological University}
\maketitle

\begin{abstract}
Adversarial training (AT) has proven to be one of the most effective ways to defend Deep Neural Networks (DNNs) against adversarial attacks. However, the phenomenon of robust overfitting, i.e., the robustness will drop sharply at a certain stage, always exists during AT. It is of great importance to decrease this robust generalization gap in order to obtain a robust model. In this paper, we present an in-depth study towards the robust overfitting from a new angle. We observe that consistency regularization, a popular technique in semi-supervised learning, has a similar goal as AT and can be used to alleviate robust overfitting. We empirically validate this observation, and find a majority of prior solutions have implicit connections to consistency regularization. Motivated by this, we introduce a new AT solution, which integrates the consistency regularization and Mean Teacher (MT) strategy into AT. Specifically, we introduce a teacher model, coming from the average weights of the student models over the training steps. Then we design a consistency loss function to make the prediction distribution of the student models over adversarial examples consistent with that of the teacher model over clean samples. Experiments show that our proposed method can effectively alleviate robust overfitting and improve the robustness of DNN models against common adversarial attacks.
\keywords{Adversarial training, robust overfitting, consistency loss}
\end{abstract}

\section{Introduction}
\label{sec1}

Recent years have witnessed the remarkable success of Deep Neural Networks (DNNs) in many artificial intelligence fields, ranging from speech recognition \cite{povey2011a}, computer vision \cite{DBLP:conf/cvpr/HeZRS16} to natural language processing \cite{devlin2018bert}. Despite their excellent performance, DNNs are vulnerable to adversarial attacks \cite{szegedy2013a,goodfellow2014a,DBLP:conf/sp/Carlini017}, which deceive the models into making wrong predictions by adding imperceptible perturbations to the natural input. Such a vulnerability can severely threaten many security-critical scenarios, and hinders the real-life adoption of DNN models.

A variety of defense methods have been proposed to improve the robustness of DNNs, e.g., feature compression \cite{xu2017feature}, input denoising \cite{guo2017countering,liao2018defense,zhang2021defense}, randomization \cite{xie2017mitigating}, gradient regularization \cite{gu2014towards}, defense distillation \cite{papernot2017a}, etc. However, most solutions were subsequently proved to be ineffective against advanced adaptive attacks \cite{athalye2018obfuscated,tramer2020adaptive}. Among these defense directions, adversarial training (AT) \cite{madry2017towards} is generally regarded as the most promising strategy. The basic idea is to craft adversarial examples (AEs) to augment the training set for model robustness improvement. A standard AT method is to use Projected Gradient Descent (PGD-AT) for AE generation \cite{madry2017towards}. Later on, more advanced approaches were designed to further enhance the robustness. For instance, TRADES \cite{zhang2019theoretically} tried to minimize the cross entropy loss for clean samples and KL divergence loss for AEs, to balance the trade-off between model robustness and natural accuracy.

\begin{wrapfigure}{r}{0.5\textwidth}
\vspace{-2em}
\includegraphics[width=1\linewidth]{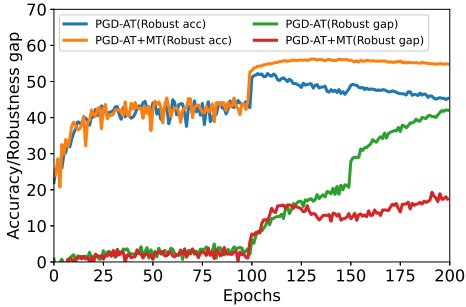}
 \vspace{-1.5em}
\caption{\small \textbf{The robust accuracy and robust generalization gap for PGD-AT without and with our MT method. We train the ResNet18 model with the CIFAR10 dataset.}}
\vspace{-2em}
\label{fig:teaser}
\end{wrapfigure}
However, AT approaches commonly suffer from one limitation: robust overfitting \cite{rice2020overfitting}. During training, the model can exhibit a gap between the robust accuracy of training set and test set, and this gap will gradually increase as the training progresses. Due to this gap, the actual model robustness can be significantly affected. Fig. \ref{fig:teaser} demonstrates this phenomenon for PGD-AT: the robust gap keeps increasing during training (green line), which can harm the model's robust accuracy on the test set (blue line). Hence, it is necessary to reduce such gap in AT.

Some attempts have been made to understand and resolve the robust overfitting phenomenon. For instance, Rice et al. \cite{rice2020overfitting} proposed to use various regularization techniques (e.g. early stopping) to alleviate robust overfitting. Chen et al. \cite{chen2020robust} integrated self-training into AT to smooth the model. Dong et al. \cite{dong2021exploring} hypothesized that robust overfitting comes from the memorization effect of the model on one-hot labels: as the one-hot labels of some samples are noisy, the model will remember those ``hard'' samples with noisy labels, leading to a decrease in robustness. Some works studied robust overfitting with weight loss landscape \cite{wu2020adversarial,Stutz2021ICCV}. However, there is still a lack of general understanding about the overfitting issue, and more effective mitigation solutions are urgently needed.
    
In this paper, we study the robust overfitting problem from a new perspective. We observe that \textit{consistency loss plays a critical role in alleviating robust overfitting in AT.} Consistency loss has been widely used in semi-supervised learning, to improve the model's confidence in predicting unlabeled data \cite{laine2016temporal,tarvainen2017mean,DBLP:conf/iclr/AthiwaratkunFIW19,berthelot2019mixmatch,berthelot2019remixmatch,xie2019unsupervised,sohn2020fixmatch}. It forces the model to give the same output distribution when the input or weights are slightly perturbed. This perfectly matches the aim of AT, which forces the model to give the same output distribution for natural or perturbed samples. Therefore, we hypothesize the integration of the consistency loss into the AT loss function can improve the model's robust generalization. We perform experiments from different aspects to validate this hypothesis. More interestingly, we review existing solutions in alleviating robust overfitting, and find most of them have implicit connections to the regularization of consistency loss.

Inspired by the above observation, we propose a new strategy to mitigate robust overfitting. It adopts ``Mean Teacher'' (MT) \cite{tarvainen2017mean}, an advanced consistency regularization method from semi-supervised learning into AT. Specifically, we introduce a teacher model during training, which comes from the average weights of the student model over different training steps. Then we adopt the consistency loss to make the student model's prediction distribution of AEs consistent with the teacher model's prediction distribution of clean samples. In this way, the trained teacher model is more robust with a smaller robust generalization gap.

Our MT-based strategy is general and can be combined with existing state-of-the-art AT solutions (e.g., PGD-AT, TRADES, etc.) to further improve the robustness and reduce the robust generalization gap. 
We comprehensively verify its effectiveness on three datasets: CIFAR10, CIFAR100 and SVHN. Taking PGD-AT as an example (Fig. \ref{fig:teaser}), our method can increase the robust accuracy of the ResNet18 model on CIFAR10 by 4\% against the PGD-10 attack, and by 3.79\% against the AutoAttack (AA) \cite{croce2020reliable}. Meanwhile, the robust generalization gap can be decreased by 24.64\%. 



\section{Background and Related Works}
\label{sec:related}

\subsection{Adversarial Training}
\label{sec2.1}
AT is a commonly used technique for learning a robust DNN model. Its basic idea is to augment the training set with adversarial examples. Formally, we aim to train the parameters $\boldsymbol{\theta}$ of a DNN model $f$ from a given training dataset of $n$ samples: $\mathcal{D}=\left\{\left(\mathbf{x}_{i},y_{i}\right)\right\}_{i=1}^{n}$, where $\mathbf{x}_{i} \in \mathbb{R}^{d}$ is a natural example and $y_{i} \in\{1, \ldots, C\}$ is its corresponding label. AT can be described as the following two-stage optimization problem: 
\begin{equation}
	\label{eq1}
	\min _{\boldsymbol{\theta}} \sum_{i=1}^{n} \max _{\mathbf{x}_{i}^{\prime} \in \mathcal{S}\left(\mathbf{x}_{i}\right)} \mathcal{L}\left(f\left(\mathbf{x}_{i}^{\prime};\boldsymbol{\theta}\right), y_{i}\right)
\end{equation}
where 
$\mathcal{L}$ is the classification loss (e.g., cross entropy), $\mathcal{S}(\mathbf{x})=\left\{\mathbf{x}^{\prime}:\left\|\mathbf{x}^{\prime}-\mathbf{x}\right\|_{p} \leq \epsilon\right\}$ is the adversarial region with $\mathbf{x}$ as the center and radius $\epsilon>0$ under the ${L}_p$ norm (e.g., ${L}_2$, ${L}_\infty$) constraint. The first stage (internal maximization optimization) is to generate the AEs for data augmentation. The second stage (external minimization optimization) is to train a robust model. 

\vspace{3pt}
\noindent\textbf{PGD-AT.}
One typical AT strategy 
is to adopt Projected Gradient Descent (PGD) \cite{madry2017towards} in the first stage, which starts from a randomly initialized point in $\mathcal{S}(\mathbf{x}_i)$ and iteratively updates it under the ${L}_\infty$ norm constraint by
\begin{equation}
	\label{eq2}
	\mathbf{x}_{i}^{\prime}=\Pi_{\mathcal{S}\left(\mathbf{x}_{i}\right)}\left(\mathbf{x}_{i}^{\prime}+\alpha \cdot \operatorname{sign}\left(\nabla_{\mathbf{x}} \mathcal{L}\left(f\left(\mathbf{x}_{i}^{\prime};\boldsymbol{\theta}\right),  y_{i}\right)\right)\right)
\end{equation}
where $\Pi(\cdot)$ is the projection operation, and $\alpha$ is the step size. Then in the second stage, it uses the generated AEs to update the model parameters $\boldsymbol{\theta}$ via the gradient descent algorithm.

\vspace{3pt}
\noindent\textbf{TRADES.}
Zhang et al. \cite{zhang2019theoretically} introduced TRADES, a new AT strategy to balance the trade-off between robustness and natural accuracy. It maximizes the Kullback–Leibler (KL) divergence between the predicted probabilities of clean samples and AEs, and then minimizes the cross-entropy loss and the adversarial loss at the same time in the second stage. The process can be formulated as
\begin{equation}
	\label{eq3}
	\min_{\boldsymbol{\theta}} \sum_{i=1}^{n}\{\mathcal{L}\left(f\left(\mathbf{x}_{i};\boldsymbol{\theta}\right), y_{i}\right) + \beta \cdot \max _{\mathbf{x}_{i}^{\prime} \in \mathcal{S}\left(\mathbf{x}_{i}\right)} \mathcal{KL}\left(f\left(\mathbf{x}_{i};\boldsymbol{\theta}\right) \| f\left(\mathbf{x}_{i}^{\prime};\boldsymbol{\theta}\right)\right)\}
\end{equation}
where $\mathcal{KL}$ is the KL divergence and $\beta$ is used to balance the trade-off.

Based on these two AT strategies, recent works about AT made many improvements, including designing new adversarial loss functions \cite{Wang2020Improving,wu2020adversarial,ding2018mma} and algorithms \cite{rice2020overfitting,pang2020bag,gowal2020uncovering}, improving training efficiency \cite{shafahi2019adversarial,wong2020fast,zhang2019you}, training with more unlabeled data \cite{uesato2019labels,carmon2019unlabeled,hendrycks2019using,zhai2019adversarially}, etc.

\subsection{Robust Overfitting in AT}
Robust overfitting is a common phenomenon in many AT solutions. During training, the model can exihibit higher robustness on the the training set than the test set. This robust generalization gap can make AT less effective in defeating adversarial attacks. 

A variety of works have investigated the causes behind this issue and tried to resolve it. 
Chen et al. \cite{chen2020robust} hypothesized that one reason of robust overfitting is that the model ``overfits'' the AEs generated in the early stage of AT and fails to generalize or adapt to the AEs in the late stage. Then they leveraged knowledge distillation and stochastic weight averaging (SWA) \cite{izmailov2018averaging} to smooth the logits and weights respectively to mitigate robust overfitting. Zhang et al. \cite{zhang2020geometry} proposed geometry-aware instance-reweighted adversarial training (GAIRAT), which assigns different weights to each sample based on the difficulty of attacking natural data points. But later this approach was shown to lead to gradient masking \cite{hitaj2021evaluating}. Huang et al. \cite{huang2020self} empirically observed that robust overfitting may be caused by the noise in the label, and proposed self-adaptive training (SAT) to soften the label and improve the generalization ability. Dong et al. \cite{dong2021exploring} explored the memorization behavior of AT and found that robust overfitting was caused by the excessive memorization of one-hot labels in typical AT methods. They proposed to integrate temporal ensemble (TE) into AT to reduce the memorization effect. Yang et al. \cite{yang2020closer} observed that the generalization ability of the model is closely related to the smoothness of local Lipschitz, i.e., a smoother local Lipschitz corresponds to better generalization ability. They proposed to use dropout to apply local Lipschitz. A number of works studied robust overfitting from the perspective of weight loss landscape \cite{li2017visualizing}, which is the loss change with respect to the weight. Wu et al. \cite{wu2020adversarial} empirically verified that a flatter weight loss landscape often leads to a smaller robust generalization gap in AT.
Stutz et al. \cite{Stutz2021ICCV} studied the relationship between robust generalization and flatness of the robust loss landscape in the weight space. 

However, the above analysis and defense methods are either not general or effective for improving the robust generalization. In this paper, we try to understand the robust overfitting problem from a different perspective -- consistency regularization. It helps us design a more effective solution to alleviate robust overfitting in AT. 

\subsection{Consistency Regularization}
\label{sec:consistency}
Consistency regularization has been widely used in the field of semi-supervised learning, which can force the model to become more confident in predicting labels on unlabeled data \cite{laine2016temporal,tarvainen2017mean,DBLP:conf/iclr/AthiwaratkunFIW19,berthelot2019mixmatch,berthelot2019remixmatch,xie2019unsupervised,sohn2020fixmatch}. It achieves this goal by encouraging the model to produce the same output distribution when its input or weights are slightly perturbed. For instance, two different augmentations of the same image should result in similar predicted probabilities. 
In semi-supervised learning, the consistency loss is usually measured by introducing a teacher model, which can be the model itself \cite{laine2016temporal} or its slightly perturbed version \cite{tarvainen2017mean}. In both cases, the student model measures the consistency of the teacher model.

Given two perturbed inputs $\mathbf{x}^{\prime}$, $\mathbf{x}^{\prime \prime}$ of a clean image $\mathbf{x}$ and the perturbed weights of student and teacher models $\boldsymbol{\theta}_{s}$, $\boldsymbol{\theta}_{t}$, consistency regularization penalizes the difference of predicted probabilities between the student model $f(\mathbf{x}^{\prime};\boldsymbol{\theta}_{s})$ and teacher model $f(\mathbf{x}^{\prime \prime};\boldsymbol{\theta}_{t})$ \cite{DBLP:conf/iclr/AthiwaratkunFIW19}. The loss term typically adopts the Mean Squared Error (MSE) or KL divergence:
\begin{equation}
\mathcal{L}_{\text {cons }}^{\text {MSE}}\left(\boldsymbol{\theta}, \mathbf{x}\right)=\left\|f\left(\mathbf{x}^{\prime} ; \boldsymbol{\theta}_{s}\right)-f\left(\mathbf{x}^{\prime \prime}, \boldsymbol{\theta}_{t}\right)\right\|^{2}
\label{eq:consistent-mse}
\end{equation}
\begin{equation}
\mathcal{L}_{\text {cons }}^{\mathrm{KL}}\left(\boldsymbol{\theta}, \mathbf{x}\right)=\mathcal{KL}\left(f\left(\mathbf{x}^{\prime} ; \boldsymbol{\theta}_{s}\right) \| f\left(\mathbf{x}^{\prime \prime}, \boldsymbol{\theta}_{t}\right)\right)
\label{eq:consistent-kl}
\end{equation}
Laine et al. \cite{laine2016temporal} introduced the $\Pi$ model, which assumes the dual roles of the teacher and student: the student learns as before; while the teacher generates goals and then learns following the student. Tarvainen et al. \cite{tarvainen2017mean} designed ``Mean Teacher'', which uses the exponential moving average of the model parameter values as the teacher model to provide a more stable target. Miyato et al. \cite{miyato2018virtual} proposed to enforce the
consistency between predictions on natural and noisy data perturbed in an adversarial direction to boost the semi-supervised learning. 

We observe that \textit{consistency regularization can perfectly match the goal of AT, which attempts to make the model output the same prediction for natural and maliciously perturbed examples.} Therefore, it offers a good opportunity to apply consistency regularization to alleviating robust overfitting. However, there are very few studies focusing on this direction. 
Tack et al. \cite{tack2021consistency} introduced a consistency regularization into AT to improve the robust generalization, which forces two AEs with different data augmentations from the same sample to produce similar prediction distributions. 
Different from \cite{tack2021consistency}, we perform a more general robustness analysis, and discover that lots of prior solutions can be abstracted as the consistency regularization. We further integrate the consistency regularization commonly used in semi-supervised learning into AT, which forces the clean output of the ensemble teacher model to be consistent with the adversarial output of the student model. This gives us higher robust generalization.

\begin{figure*}
	\centering
	\subfigure[]{
		\includegraphics[width=.3\linewidth]{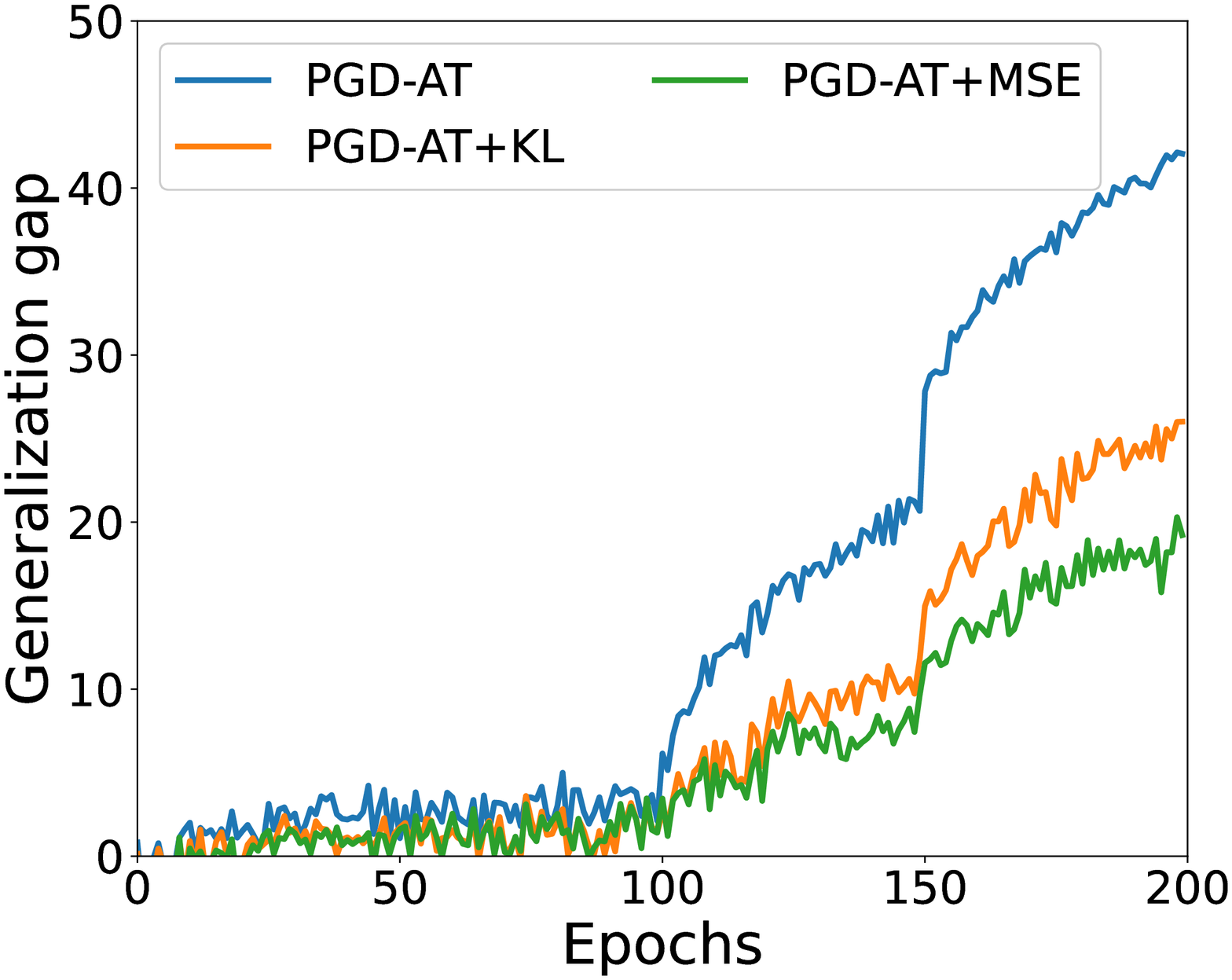}
		\label{fig1_a}
	}
	\subfigure[]{
		\includegraphics[width=.3\linewidth]{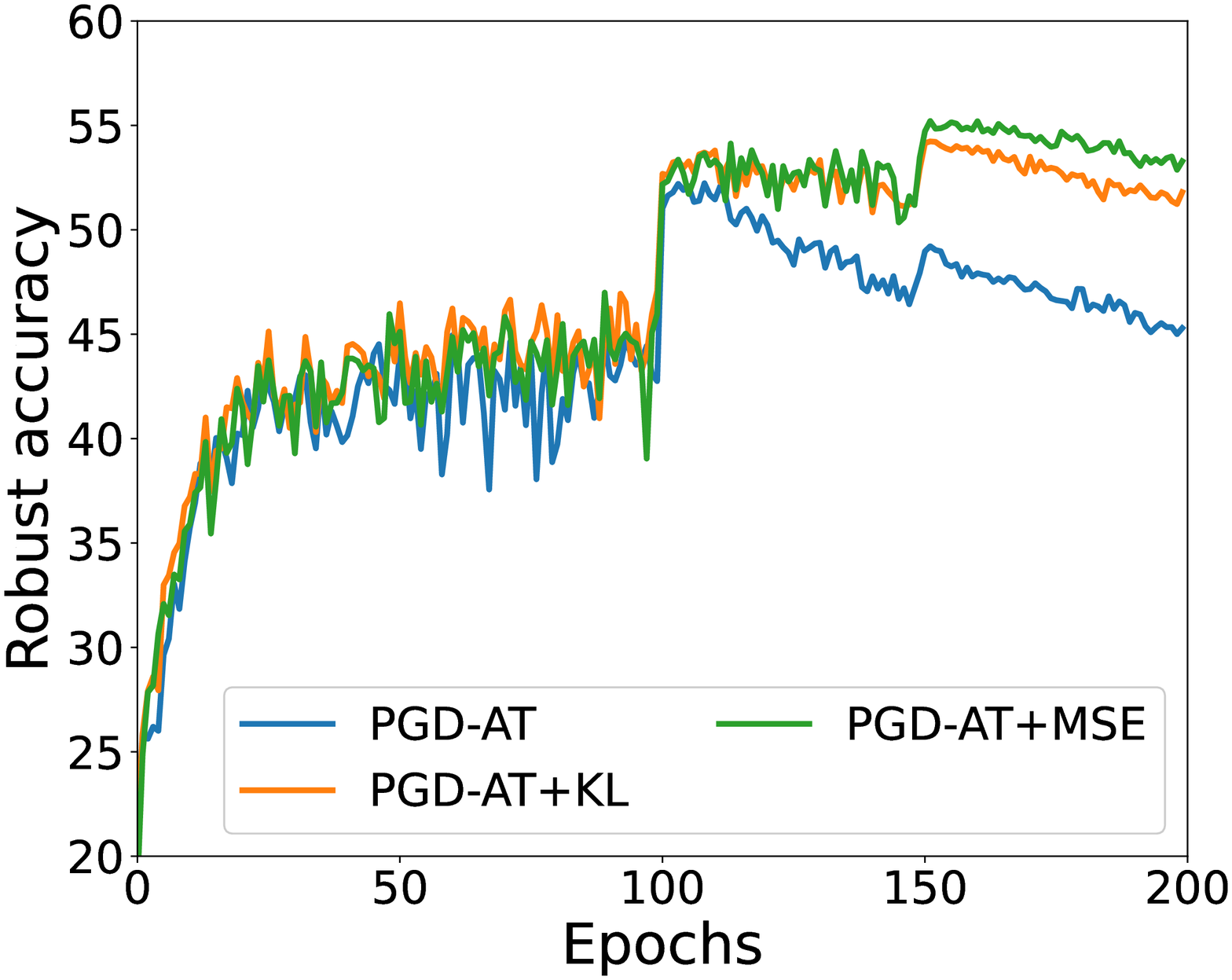}
		\label{fig1_b}
	}
	\subfigure[]{
		\includegraphics[width=.3\linewidth]{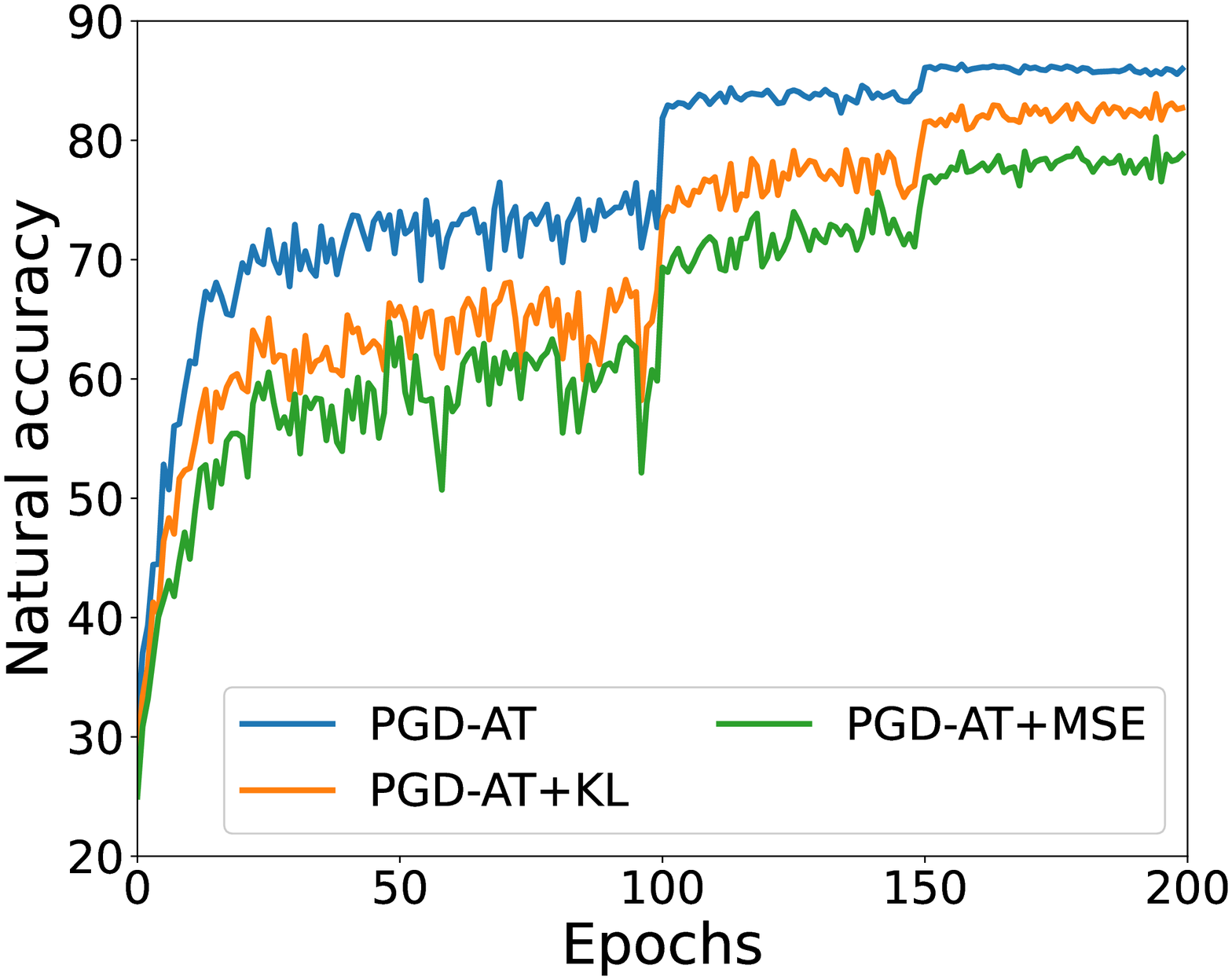}
		\label{fig1_c}
	}
	\vspace{-1em}
	\caption{\small \textbf{(a) Robust generalization gap between the training and test sets; (b) Robust accuracy of the test set; (c) Natural accuracy of the test set.}}
	\label{fig1}
\end{figure*}
\section{Robust Overfitting with Consistency Regularization}
\label{sec3}

As discussed in Section \ref{sec:consistency}, consistency regularization might be a promising strategy to alleviate the robust overfitting issue. In this section, we perform an in-depth analysis to confirm its effectiveness. We first empirically disclose the impact of consistency regularization in AT from different perspectives (Section \ref{sec3.1}). Then we revisit existing solutions and find they can be abstracted as consistency regularization (Section \ref{sec3.2}).



\subsection{AT With Consistency Loss}
\label{sec3.1}

We perform comprehensive experiments to show the impact of consistency regularization in reducing the robustness generalization gap. 
We consider three AT methods. (1) PGD-AT: we adopt the PGD-10 attack with the step size $\alpha=2/255$ and the maximum perturbation bound $\epsilon=8/255$ under the $L_{\infty}$ norm. (2) PGD-AT+KL: we integrate the consistency regularization of KL divergence (Equation \ref{eq:consistent-kl}) to the loss function of PGD-AT. We use the model itself as the teacher model to measure the consistency loss of prediction probabilities between natural and adversarial examples. (3) PGD-AT+MSE: similar as PGD-AT+KL, we include the consistency regularization of MSE (Equation \ref{eq:consistent-mse}). 
For all the three approaches, we train the ResNet18 \cite{DBLP:conf/cvpr/HeZRS16} model on CIFAR10 for 200 epochs. The initial learning rate is 0.1, and decays by a factor of 10 at the 100\textit{th} and 150\textit{th} epochs.

\vspace{3pt}
\noindent\textbf{Training and Test Robustness}.
We measure the robustness of the training and test sets, which is calculated as the prediction accuracy of AEs crafted from these two sets. We have the following observations. First, \textit{the adoption of consistency loss can help reduce robust overfitting}. Fig. \ref{fig1_a} shows the robust generalization gap between the training and test set at different training epochs, which quantifies the robust overfitting degree. We observe that in the first 100 epochs, the robust generalization gap for all these methods is close to zero. After the first learning rate decay, the gap of these solutions increases with the training process. PGD-AT+KL and PGD-AT+MSE has much smaller gap than PGD-AT, attributed to the regularization of consistency loss. 

Second, \textit{consistency loss based on MSE has stronger regularization effect and is more effective against robust overfitting than the KL divergence.} In Fig. \ref{fig1_a}, we can see PGD-AT+MSE has smaller robust generalization gap than PGD-AT+KL. Fig. \ref{fig1_b} shows the robust accuracy over the test set for different approaches. We also observe PGD-AT+MSE has higher test robustness than the other two, which results in smaller robust generalization gap. 

Third, \textit{strong regularization can also cause a decrease in natural accuracy.} Fig. \ref{fig1_c} compares the natural accuracy of different solutions. We find PGD-AT+MSE has the lowest accuracy on natural samples. This indicates a trade-off between natural accuracy and robust generalization gap.

\vspace{3pt}



\begin{wrapfigure}{r}{0.5\textwidth}
\vspace{-2em}
\includegraphics[width=1\linewidth]{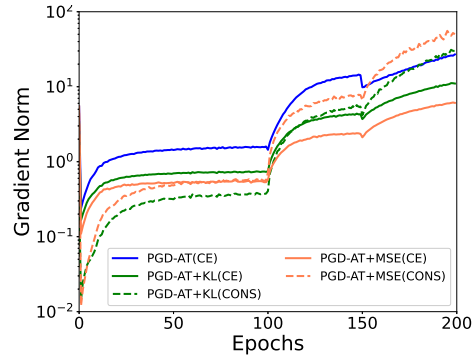}
 \vspace{-15pt}
\caption{\small \textbf{Gradient norms of the cross-entropy term (CE) and consistency loss term (CONS).}}
\vspace{-10pt}
\label{fig1_d}
\end{wrapfigure}
\noindent\textbf{Average Gradient Norms of Loss Terms.}
To further understand the importance of consistency regularization, we compute the average gradient norms of the cross-entropy (CE) term $\left\|\nabla \mathcal{L}_{\mathrm{ce}}\right\|$ and consistency loss (CONS) term $\left\|\nabla \mathcal{L}_{\mathrm{cons}}\right\|$ in the adversarial loss, respectively. Fig. \ref{fig1_d} shows the trends of these two metrics for different approaches. First, we observe that in the initial training stage, $\left\|\nabla \mathcal{L}_{\mathrm{ce}}\right\|$ in PGD-AT+MSE and PGD-AT+KL is larger than $\left\|\nabla \mathcal{L}_{\mathrm{cons}}\right\|$, indicating that the CE loss dominants in the early stage. As the training proceeds, $\left\|\nabla \mathcal{L}_{\mathrm{cons}}\right\|$ gradually increases and dominates the training after the 100\textit{th} epoch. Second, in PGD-AT, $\left\|\nabla \mathcal{L}_{\mathrm{ce}}\right\|$ continues to increase and always remains at a higher value, which leads to the overfitting in the later stage. In contrast, due to the constraint of consistency loss, PGD-AT+KL and PGD-AT+MSE can alleviate overfitting to certain extent. 

\vspace{3pt}
\noindent\textbf{Weight Loss Landscape}.
We further adopt the weight loss landscape \cite{li2017visualizing} to explore the relationship between the consistency regularization and robust overfitting.
Weight loss landscape is a very good visualization method to characterize the generalization ability of a DNN model, and has been empirically and theoretically investigated \cite{DBLP:conf/icml/DinhPBB17,DBLP:conf/iclr/ForetKMN21,DBLP:conf/iclr/JiangNMKB20,wu2020adversarial,DBLP:journals/corr/abs-2102-02950,DBLP:conf/nips/LiuSLTS20}. Its basic idea is to add some noise with  random directions and magnitudes to the model weights, and calculate the loss for each training sample to observe the loss change. The model with a strong generalization ability is not sensitive to small-scale noise, so the loss change is smaller with a flatter loss landscape, and vice versa.

For adversarially trained models, we follow the setting in \cite{wu2020adversarial} to demonstrate the weight loss landscape by plotting the adversarial loss changes along a random Gaussian direction $d$ with the magnitude $\alpha$:
\begin{equation}
	\label{eq5}
	L(\theta) = \sum_{i=1}^{n} \max _{\mathbf{x}_{i}^{\prime} \in \mathcal{S}\left(\mathbf{x}_{i}\right)} \ell \left(f\left(\mathbf{x}_{i}^{\prime};\boldsymbol{\theta}+\alpha d\right), y_{i}\right)
\end{equation}
where $\mathbf{x}_{i}^{\prime}$ is generated on-the-fly by the PGD-10 attack (10 steps PGD attack with step size $\alpha=2/255$) for the perturbed model $\boldsymbol{\theta}+\alpha d$, and $L$ refers to the CE loss.
Then we normalize each filter in $d$ to have the same norm of the corresponding filter in $\theta$. Specifically, we make the replacement $d_{i, j} \leftarrow \frac{d_{i, j}}{\left\|d_{i, j}\right\|}\left\|\theta_{i, j}\right\|$, where $d_{i,j}$ represents the $j$th filter (not the $j$th weight) of the $i$th layer of $d$, and $\|\cdot\|$ denotes the Frobenius norm. 

\begin{figure}[t]
	\centering
	\subfigure[``Best'' model.]{
		\includegraphics[width=0.47\linewidth]{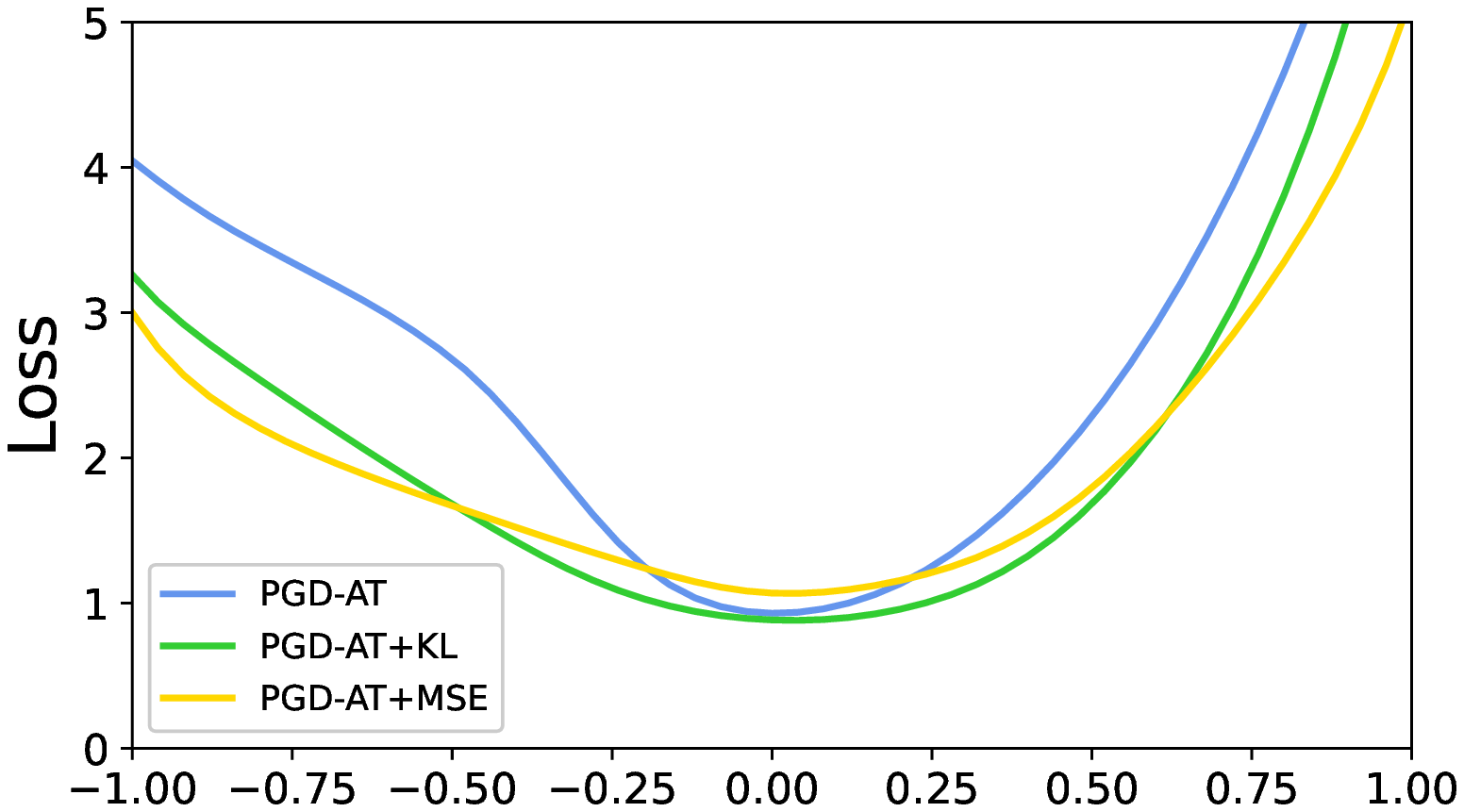}
		\label{fig2_a}
	}
	\subfigure[``Last'' model.]{
		\includegraphics[width=0.47\linewidth]{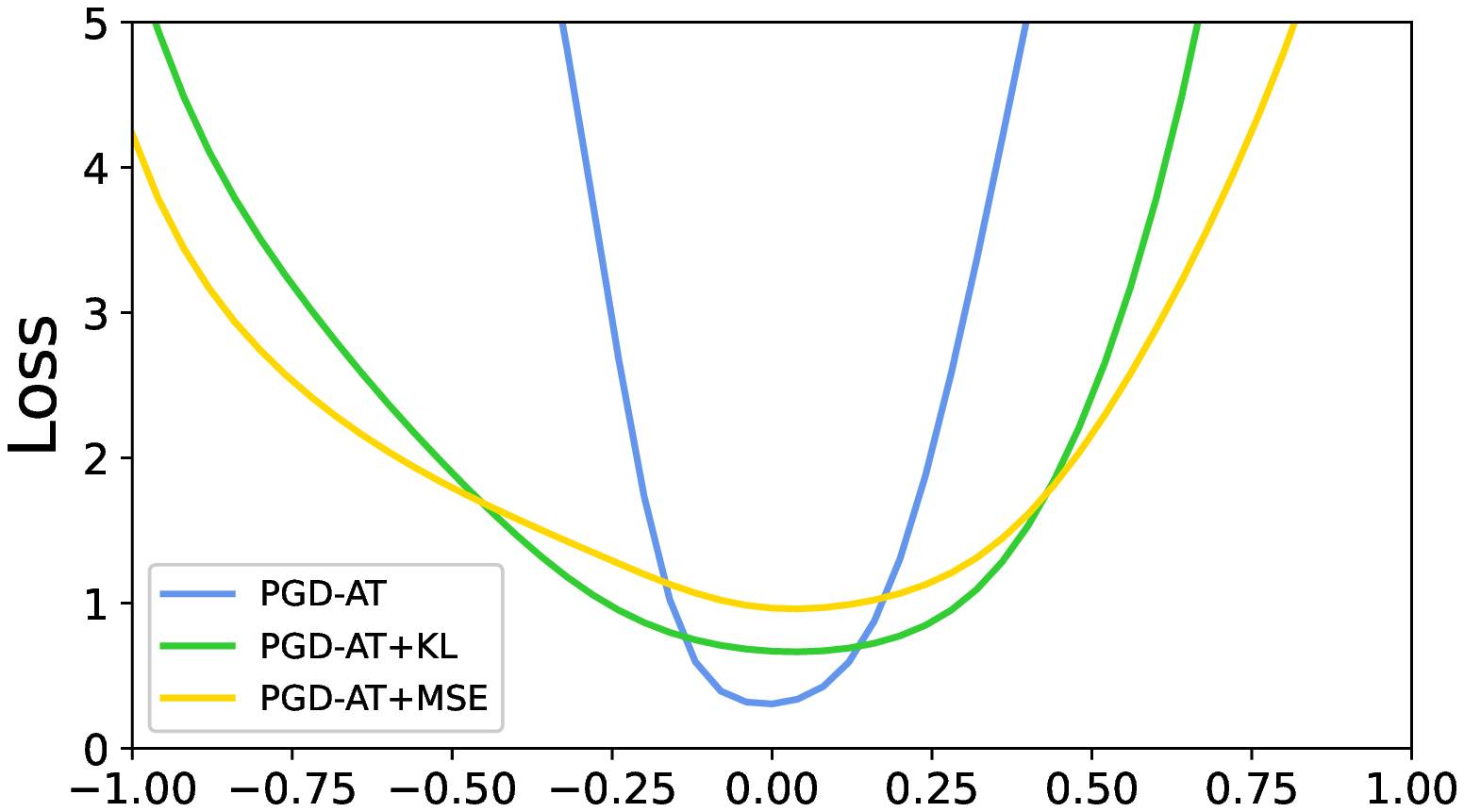}
		\label{fig2_b}
	}
	\vspace{-1em}
	\caption{\small \textbf{Adversarial weight loss landscape for various methods.}}
	\vspace{-1em}
	\label{fig2}
\end{figure}
Fig. \ref{fig2} shows the results of different AT approaches. We draw the adversarial weight loss landscape of the ``Best'' (the checkpoint with the highest test robustness under the PGD-10 attack) and ``Last'' models (the last checkpoint of training). We observe that for the ``Best'' model (Fig. \ref{fig2_a}), the weight loss landscapes are relatively flat. For the ``Last'' model (Fig. \ref{fig2_b}), the adversarial weight loss landscapes are sharper. Comparatively, PGD-AT+MSE and PGD-AT+KL have flatter weight loss landscapes than PGD-AT, as they have the consistency loss. This confirms the consistency regularization can reduce the robust generalization gap and lead to flatter weight loss landscape.

\subsection{Revisiting Existing Solutions for Robust Overfitting}
\label{sec3.2}

According to the above empirical analysis, consistency regularization can force the model to have similar predicted probability distributions for natural and adversarial examples, avoid overconfidence in model predictions, and implicitly flatten the weight loss landscape. Hence, we believe it can effectively mitigate the robust overfitting problem. We look through prior solutions that aimed to address the robust overfitting issue. Interestingly, we find most solutions can be implicitly connected to different forms of regularization consistency, even the authors did not realize or mention the significant impact of consistency loss. 

Particularly, (1) different works may choose different ``teacher models'' for regularizing the consistency loss, such as the model itself \cite{zhang2019theoretically,tack2021consistency}, the ensemble predictions of the previous epoch \cite{huang2020self,dong2021exploring}, or other pre-trained models \cite{chen2020robust}. (2) The consistency regularization can be computed by different metrics, such as MSE \cite{chen2020robust,huang2020self,dong2021exploring}, KL \cite{zhang2019theoretically}, and Jensen-Shannon \cite{tack2021consistency}). (3) The loss can also compute the distance of different types of samples, including adversarial-natural samples \cite{zhang2019theoretically,huang2020self,dong2021exploring}, adversarial-adversarial samples \cite{chen2020robust,tack2021consistency} and natural-natural samples \cite{chen2020robust}). We conclude that \textit{the consistency regularization is a fundamental mechanism behind many solutions to address the robust overfitting problem. }

\section{A New Solution Based on Consistency Regularization}
\label{sec4}

From Section \ref{sec3.2} we find that existing solutions implicitly adopted the idea of consistency loss to achieve good robust generalization performance. We hypothesize that using more advanced consistency regularization approaches could further alleviate robust overfitting. Therefore, we propose a new defense methodology, which leverages the Mean Teacher (MT) strategy \cite{tarvainen2017mean} in semi-supervised learning to enforce the consistency loss during AT. Our method can be integrated with existing AT solutions, e.g., PGD-AT, TRADES, RST \cite{carmon2019unlabeled}, MART \cite{Wang2020Improving}, etc. Evaluations in Section \ref{sec5} show our MT-based AT can outperform prior solutions that are implicitly based on consistency regularization. 

Specifically in Section \ref{sec3.1}, we use the trained student model itself as the teacher model to minimize the probability distribution of natural and adversarial examples. However, this strategy is not optimal, because it cannot fully utilize the ``dark knowledge'' \cite{hinton2014dark}, and will result in a decrease in the natural accuracy (Fig. \ref{fig1_c}). 
To overcome such limitation, our MT-based AT builds a teacher model from the Exponential Moving Average (EMA) weights of an ensemble of student models, and then adopts it to guide the consistency regularization  during training, which can improve the model's robust generalization.

We use PGD-AT to illustrate the details of our methodology. Other AT strategies can be applied as well. Particularly, for a natural sample $\mathbf{x}_i$, we obtain its corresponding adversarial example $\mathbf{x}_{i}^{\prime}$. We calculate the consistency loss between the predicted probability of the teacher model $\theta_t$ on the clean sample $\mathbf{x}_i$ and the predicted probability of the student model $\theta_s$ on the  adversarial example $\mathbf{x}_{i}^{\prime}$. Therefore, the training objective of PGD-AT with MT (PGD-AT+MT) can be expressed as: 
\begin{equation}
	\label{eq6}
	\min _{\boldsymbol{\theta}_{s}} \sum_{i=1}^{n} \max _{\mathbf{x}_{i}^{\prime} \in \mathcal{S}\left(\mathbf{x}^a_{i}\right)} \{\mathcal{L}\left(f\left(\mathbf{x}_{i}^{\prime};\boldsymbol{\theta}_{s}\right), y_{i}\right)
	+ \lambda \cdot \mathcal{L}_{\text {cons }}\left(f\left(\mathbf{x}_{i}^{\prime};\boldsymbol{\theta}_{s}\right), f\left(\mathbf{x}_{i},\boldsymbol{\theta}_{t}\right)\right)\}
\end{equation}
The weight of the teacher model $\theta_{t}$ is computed from the EMA of the student model's weights $\theta_{s}$:
\begin{equation}
	\label{eq7}
	\boldsymbol{\theta_{t}} = \eta\cdot\boldsymbol{\theta}_{t} + (1-\eta)\cdot\boldsymbol{\theta}_{s}
\end{equation}
where $\eta$ is a smoothing coefficient hyperparameter.

The overall training process of our PGD-AT+MT is described in Algorithm \ref{alg:algorithm1}. In the early training stage (first $E_s$ epochs), it is nonsense to calculate the consistency loss, as the prediction of the student model is not accurate. So we adopt the normal PGD-AT loss (Lines \ref{line:1} and \ref{line:3}). After the first learning rate decay, the model memorizes most of the internal representation of the input samples, and the robust generalization gap starts to increase (Fig. \ref{fig2_a}). So we integrate the consistency loss with MT to guide the student model and alleviate robust overiftting (Lines \ref{line:2} and \ref{line:4}).
\begin{algorithm}
    \scriptsize
	\caption{PGD-AT+MT}
	\label{alg:algorithm1}
	\begin{algorithmic}[1]
	    \REQUIRE training set $\mathcal{D}$; batch size $m$; learning rate $lr$; number of training epochs $T$; PGD step size $\eta_{1}$; number of PGD steps $K$; max perturbation budget $\epsilon$; EMA smoothing parameter $\eta$; 
	    EMA start epoch $E_{s}$
	    \ENSURE Robust model weight $\theta_{t}$
	    
	    \STATE Randomly initialize $\theta_s$
		\FOR{$t=1$ {\bfseries to} $T$}
		\FOR{$i=1$ {\bfseries to} $m$}
		\STATE Sample $\mathbf{x}_{i}$ from $\mathcal{D}$
		\STATE $\mathbf{x}_{i}^{\prime} \leftarrow \mathbf{x}_{i}+\epsilon \delta$, where $\delta \sim$ Uniform$(-1,1)$
		\FOR {$k=1$ {\bfseries to} $J$}
		\IF {$t < E_{s}$}
		\STATE $\mathbf{x}_{i}^{\prime} \leftarrow \Pi_{\epsilon}\left(\mathbf{x}_{i}^{\prime}+\eta_{1} \operatorname{sign}\left(\nabla_{\mathbf{x}_{i}^{\prime}} \mathcal{L}_{\text {ce }}\left(f\left(\mathbf{x}_{i}^{\prime};\boldsymbol{\theta}_{s}\right), y_{i}\right)\right)\right)$\label{line:1}
		\ELSE
		\STATE $\mathbf{x}_{i}^{\prime} \leftarrow \Pi_{\epsilon} (\mathbf{x}_{i}^{\prime}+\eta_{1}\operatorname{sign}(\nabla_{\mathbf{x}_{i}^{\prime}}\mathcal{L}_{\text {ce}}({f}(\mathbf{x}_{i}^{\prime};\boldsymbol{\theta}_{s}),y_{i})+ \lambda \mathcal{L}_{\text {cons }}(f(\mathbf{x}_{i};\boldsymbol{\theta}_{t}),f(\mathbf{x}_{i}^{\prime};\boldsymbol{\theta}_{s}))))$\label{line:2}
		\ENDIF
		\ENDFOR
		\IF {$t < E_{s}$}
		\STATE $\theta_{s} \leftarrow \theta_{s} - lr \cdot \nabla_{\mathbf{x}_{i}^{\prime}}(\mathcal{L}_{\text {ce }}({f}(\mathbf{x}_{i}^{\prime};\boldsymbol{\theta}_{s}),y_{i})$\label{line:3}
		\STATE $\theta_{t} = \theta_{s}$
		\ELSE
		\STATE $\theta_{s} \leftarrow \theta_{s} - lr \cdot \nabla_{\mathbf{x}_{i}^{\prime}}(\mathcal{L}_{\text {ce }}({f}(\mathbf{x}_{i}^{\prime};\boldsymbol{\theta}_{s}),y_{i})+\lambda \mathcal{L}_{\text {cons }}(f(\mathbf{x}_{i};\boldsymbol{\theta}_{t}),f(\mathbf{x}_{i}^{\prime};\boldsymbol{\theta}_{s})))$\label{line:4}
		\STATE $\theta_{t} = \eta\cdot\theta_{t} + (1-\eta)\cdot\theta_{s}$
		\ENDIF
		\ENDFOR
		\ENDFOR
		
	\end{algorithmic}
\end{algorithm}
\section{Evaluation}
\label{sec5}
We construct extensive experiments to demonstrate the effectiveness of our MT-based AT in alleviating robust overfitting and improve the model robustness. 
\subsection{Experimental Setups}
\noindent\textbf{Datasets and models.}
Our solution is general for image classification tasks with different datasets and models. Without loss of generality, we choose three common image datasets: CIFAR10, CIFAR100 \cite{krizhevsky2009learning} and SVHN \cite{netzer2011reading}. All the natural images are normalized into the range of [0, 1]. We apply simple data augmentations, including 4-pixel padding with $32 \times 32$ random crop and random horizontal flip. We mainly adopt the ResNet18 model for evaluation. We also show our method is scalable for large-scale models (WideResNet-28-10 \cite{DBLP:conf/bmvc/ZagoruykoK16}).
\vspace{3pt}
\noindent\textbf{Baselines and training details.}
Our solution can be integrated with existing AT strategies. We mainly consider two popular approaches (PGD-AT and TRADES) as the baselines and combine MT for comparisons. We perform adversarial training under the most commonly used $L_{\infty}$ norm with a maximum perturbation budget $\epsilon = 8/255$.  During training, we use the 10-step PGD attack with a step size $\alpha=2/255$ to optimize the internal maximization. We adopt the SGD optimizer with a momentum of 0.9, weight decay of 0.0005 and train batch size of 128. For WideResNet-28-10, we adopt the dropout regularization \cite{DBLP:journals/jmlr/SrivastavaHKSS14} with a drop rate of 0.3. For CIFAR10 and CIFAR100, we train 200 epochs. The initial learning rate is 0.1, and decayed by 10 at the 100\textit{th} and 150\textit{th} epoch. For SVHN, we train 80 epochs. The initial learning rate is 0,01, and decayed by the cosine annealing schedule. 

We use MSE (Equation \ref{eq:consistent-mse}) to calculate the consistency loss since it performs better than KL divergence (Section \ref{sec3.1}). When integrating our method into PGD-AT or TRADES, the MT consistency regularization is applied when the learning rate decays for the first time. We set the EMA hyperparameter $\eta=0.999$ and the consistency weight $\lambda=30$ ($\lambda=10$ for WideResNet-28-10) along a Gaussian ramp-up curve.

A few prior solutions also adopt 
some regularization methods in the loss function to improve the model robustness and alleviate robust overfitting. We choose them as baselines for comparison as well. Specifically, (1) KD+SWA \cite{chen2020robust} leverages knowledge distillation and SWA to injecting smoothness into the model weights.
(2) PGD-AT+TE/TRADES+TE \cite{dong2021exploring} apply temporal ensemble (TE) into AT to soften the target label. (3) PGD-AT+CONS/TRADES+CONS \cite{tack2021consistency} force the predictive distributions after attacking from two different augmentations of the same instance to be similar with each other. 

\vspace{3pt}
\noindent\textbf{Evaluated attacks and metrics.}
We consider three popular adversarial attacks to avoid false robustness caused by gradient confusion \cite{athalye2018obfuscated} under the white-box setting: PGD \cite{madry2017towards} with different steps (PGD-10, PGD-100 with the step size $\epsilon/4$), C\&W$_{\infty}$ (update perturbation with PGD) \cite{DBLP:conf/sp/Carlini017} and AA \cite{croce2020reliable}. Note that AA is an attack suite which also includes a black-box attack. We adopt different evaluation metrics: (1) the model accuracy over the natural samples. This is to measure the model usability. (2) The highest model accuracy over the AEs achieved on different checkpoints during training. This is to measure the model robustness. (3) The robust accuracy difference between the training set and test set under the PGD-10 attack at the last checkpoint. This is to measure the robust overfitting. An effective method should have high natural accuracy and robust accuracy, with small robust generalization gap.  

\subsection{Effectiveness and Comparisons}
Table \ref{tab1} shows the impact of MT on the adversarial training process. We have the following observations. First, the model with the MT strategy can maintain similar natural accuracy as the original approach. Second, MT can greatly enhance the model robustness against different attacks for most cases. Third, MT can also largely reduce the robust generalization gap and alleviate the robust overfitting. Table \ref{tab2} further shows the impact of MT on the WideResNet-28-10 model with the CIFAR10 dataset. It implies our MT-based strategy is effective for large CNN models as well. 

\begin{table}[t]
    \centering
	\caption{The impact of MT on the different datasets and AT approaches.}
	\label{tab1}
	\resizebox{\textwidth}{!}{
	\begin{tabular}{c|c|c|cccc|c}
	
		\Xhline{1pt}
		\multirow{2}{*}{Dataset}                 &  \multirow{2}{*}{Training Strategy}    & Natural 			& \multicolumn{4}{c|}{Robust Accuracy} & Generalization 		  \\ \cline{4-7}
		&  & Accuracy & PGD-10 & PGD-100 & C\&W-100 & AA & Gap\\\Xhline{1pt}
		\multirow{4}{*}{CIFAR10} & PGD-AT   & \textbf{83.62}  	& 52.25  		  & 50.79  			& 49.47 		 & 46.96 		  & 42.05    	  \\
							 	 & PGD-AT+MT (Ours) & 83.56   			& \textbf{56.25}  & \textbf{55.71}  & \textbf{52.30} & \textbf{50.75} 		  & \textbf{17.41}    	  \\ \cline{2-8} 
							 	 & TRADES    & \textbf{82.69}   			& 53.87			  & 53.28	 		& 50.69 		 & 49.57 		  & 26.17    	  \\
							 	 & TRADES+MT (Ours) & 82.36   			& \textbf{55.44}  		  & \textbf{54.69}  			& \textbf{52.19} 		 & \textbf{51.16} & \textbf{15.50} \\ \Xhline{1pt}

		\multirow{4}{*}{CIFAR100} & PGD-AT   & 56.96   		& 28.78 		 & 28.37 		  & 26.75 & 24.13 			& 69.05 \\
						  & PGD-AT+MT (Ours) 		  & \textbf{58.17}   		& \textbf{30.22}  		 & \textbf{29.81}  		  & \textbf{27.97}  		   & \textbf{26.03} & \textbf{36.11} \\ \cline{2-8} 
						  & TRADES   		  & 57.14   		& 29.88  		 & 29.60  		  & 25.29  		   & 24.20 			& 52.30 \\
						  & TRADES+MT (Ours) 		  & \textbf{58.20}  & \textbf{30.69} & \textbf{30.44} & \textbf{26.59} 		   & \textbf{25.72} 			& \textbf{32.86} \\ \Xhline{1pt}
		
		\multirow{4}{*}{SVHN} & PGD-AT   & 88.96   	   & 54.36  		& 53.24  		 & 48.11 		  & 46.41 		   & 48.59    \\
					  		  & PGD-AT+MT (Ours) & \textbf{91.67} & \textbf{60.61}  		& \textbf{59.06}  		 & \textbf{53.93} 		  & \textbf{50.71} & \textbf{26.53}    \\ \cline{2-8} 
			      			  & TRADES   & \textbf{90.56}   	   & \textbf{62.52} & 60.63  		 & \textbf{56.61} & 43.20 		   & 30.02    \\
					  		  & TRADES+MT (Ours) & 90.48   	   & 61.94  		& \textbf{60.69} & 54.90 		  & \textbf{50.68} 		   & \textbf{23.37}    \\ \Xhline{1pt}
	\end{tabular}
    }
\end{table}

\begin{table*}[t]
	\centering
	\caption{The impact of MT on the WideResNet-28-10 model.}
	\label{tab2}
	\resizebox{\textwidth}{!}{
	\begin{tabular}{c|c|c|cccc|c}
			\Xhline{1pt}
			\multirow{2}{*}{Dataset} & \multirow{2}{*}{Training Strategy}      & Natural 		   	& \multicolumn{4}{c|}{Robust Accuracy} & Generalization \\ \cline{4-7}
			&  & Accuracy & PGD-10 & PGD-100 & C\&W-100 & AA & Gap\\\Xhline{1pt}
			\multirow{4}{*}{CIFAR10} & PGD-AT     & \textbf{86.84}   		   & 55.62  		& 54.76   		  & 54.16    	   & 51.56	    & 50.82 \\
			& PGD-AT+MT (Ours)   & 84.52 	   & \textbf{58.38} & \textbf{57.86}  & \textbf{56.19}    	   & \textbf{53.46}          & \textbf{21.16} \\ \cline{2-8}
			& TRADES	& 83.96  		   & 56.46			& 55.95		      &	53.24    	   & 52.14		    & 43.99 \\
			& TRADES+MT (Ours)	& \textbf{85.28}   		   & \textbf{58.93}	& \textbf{58.33}  &	\textbf{56.28} & \textbf{55.03} & \textbf{18.06} \\ \Xhline{1pt}
	\end{tabular}%
	}
\end{table*}

\begin{table*}[t]
	\centering
	\caption{Comparison of MT with other related works.}
	\label{tab3}
	\resizebox{0.9\textwidth}{!}{
	\begin{tabular}{c|c|cccc|c}
			\Xhline{1pt}
			\multirow{2}{*}{Training strategy}       & Natural 		   	& \multicolumn{4}{c|}{Robust Accuracy} & Generalization \\ \cline{3-6}
			& Accuracy & PGD-10 & PGD-100 & C\&W-100 & AA & Gap\\\Xhline{1pt}
			PGD-AT     & 83.62   		   & 52.25  		& 50.79   		  & 49.47    	   & 46.96 		    & 42.05 \\
			KD+SWA \cite{chen2020robust}      & 83.63   		   & 53.56  		& 52.97   		  & 51.23          & 49.22 		    & 20.77 \\
			PGD-AT+TE \cite{dong2021exploring}   & 82.56   		   & 56.03  		& 55.38   		  & \textbf{52.50} & 50.30 		    & 18.76 \\
			PGD-AT+CONS \cite{tack2021consistency} & \textbf{84.73}   & 55.64          & 54.90   		  & 51.58    	   & 49.16 		    & 17.64 \\
			PGD-AT+MT (Ours)   & 83.56   	   	   & \textbf{56.25} & \textbf{55.71}  & 52.30    	   & \textbf{50.75} & \textbf{17.41} \\ \Xhline{1pt}
			TRADES	& 82.69  		   & 53.87			& 53.28		      &	50.69    	   & 49.57		    & 26.17 \\
			TRADES+TE \cite{dong2021exploring}	& \textbf{83.88}   & 54.65			& 53.89			  & 50.40	 	   & 48.97		    & 20.30 \\
			TRADES+CONS \cite{tack2021consistency}	& 82.94   		   & 55.07			& 54.38			  &	50.49 		   & 49.34	        & \textbf{12.25} \\
			TRADES+MT (Ours)	& 82.36   		   & \textbf{55.44}	& \textbf{54.69}  &	\textbf{52.19} & \textbf{51.16} & 15.50 \\ \Xhline{1pt}
	\end{tabular}%
	}
\end{table*}

Table \ref{tab3} presents the comparisons of our work with other related ones. All the models are trained on CIFAR10. These state-of-the-art methods can effectively reduce the robust generalization gap, and improve the model robustness. Comparatively, our solution has the highest robustness against most adversarial attacks. This demonstrates the superiority of MT when integrated with AT.

\subsection{Ablation Studies}
We perform ablation studies to disclose the impact of each component in our proposed method. Without loss of generality, we choose the PGD-AT+MT as the training strategy, and CIFAR10 dataset. We use the PGD-10 attack to evaluate the model robustness.
\vspace{3pt}

\begin{wrapfigure}{r}{0.6\textwidth}
\vspace{-30pt}
  \centering
    \subfigure[]{
		\includegraphics[width=.45\linewidth]{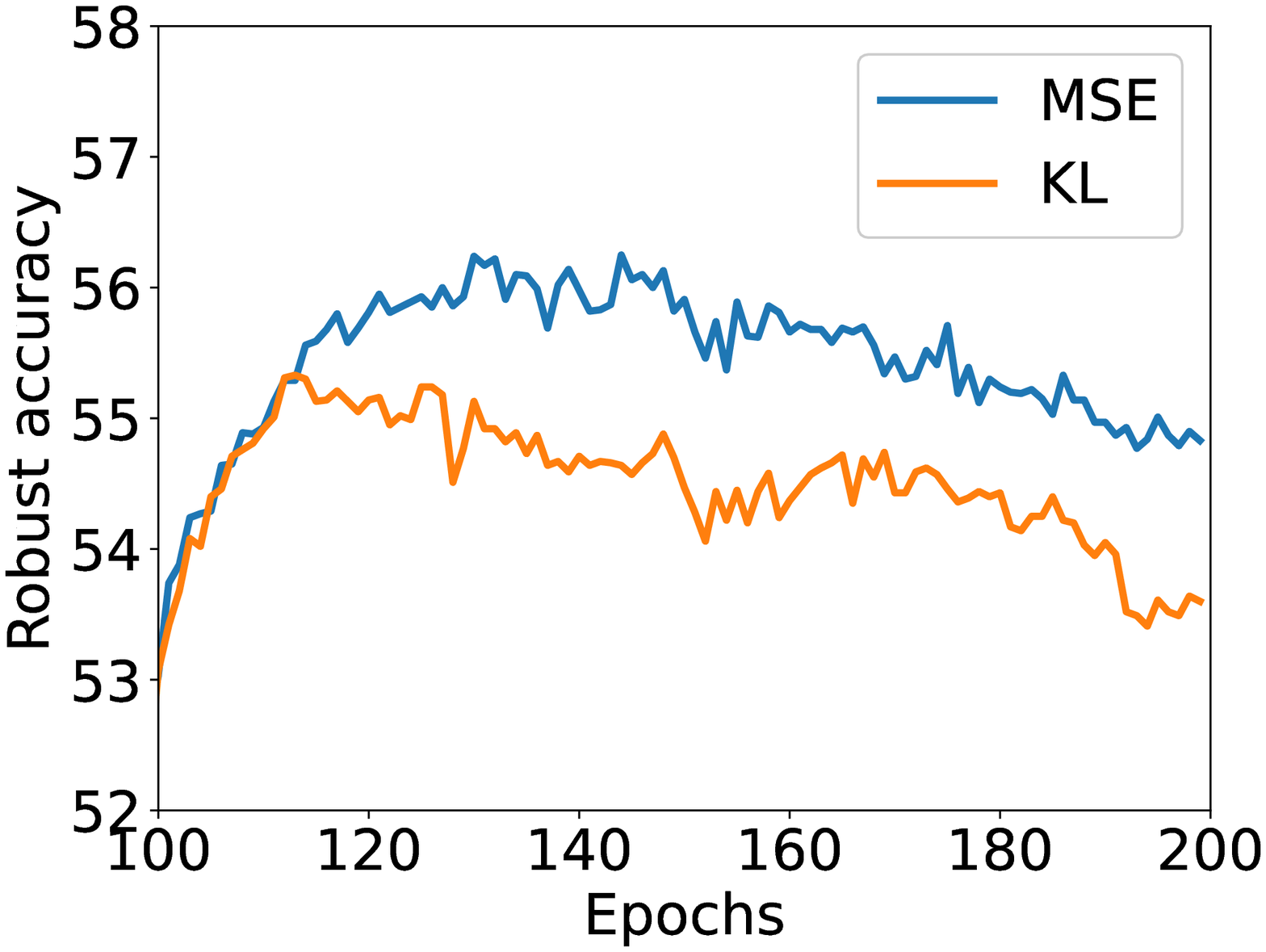}
		\label{fig4_a}
	}
    \subfigure[]{
	    \includegraphics[width=.45\linewidth]{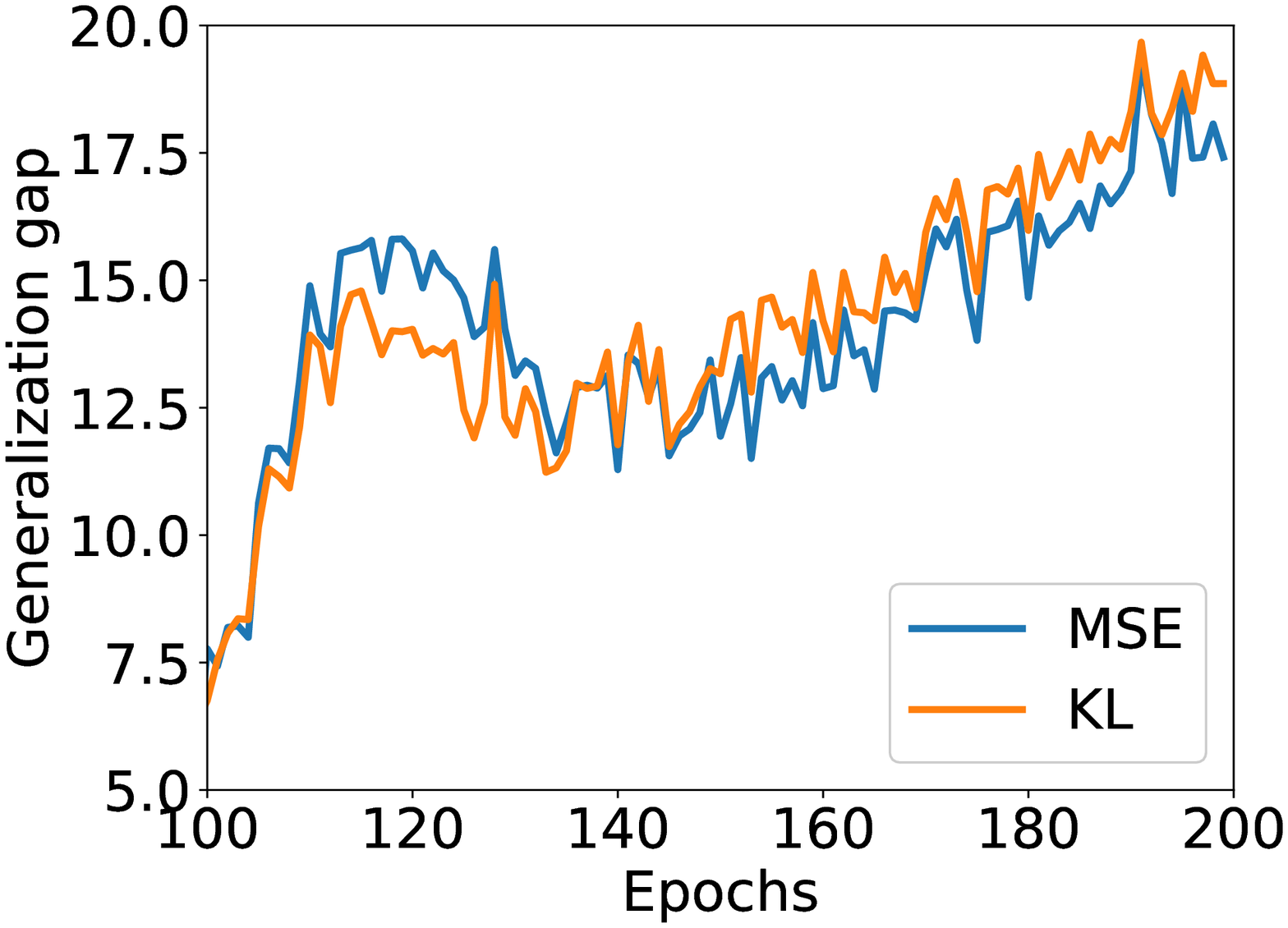}
	    \label{fig4_b}
	    }
	\vspace{-1em}
	\caption{\small \textbf{Comparisons of consistency loss. (a) Test robust; (b) Robust generalization gap}}
	\vspace{-20pt}
	\label{fig4}
 \end{wrapfigure}
\noindent\textbf{Impact of consistency loss.} 
As discussed in Section \ref{sec:consistency}, there are mainly two types of consistency loss: MSE (Equation \ref{eq:consistent-mse}) and KL divergence (Equation \ref{eq:consistent-kl}). We measure their performance in reducing the robust overfitting. For fair comparisons, we restore the model checkpoint from the 99\textit{th} epoch and then apply the two loss functions respectively. All the other configurations and hyper-parameters are the same. Fig. \ref{fig4} shows the test robustness and robust generalization gap for each loss. We observe that the MSE loss always has slightly higher test robustness, and lower robust generalization gap compared to KL divergence. One possible reason is that MSE can better alleviate the model output overconfident predictions, which could be the cause of robust overfitting \cite{dong2021exploring}.

\begin{wrapfigure}{r}{0.6\textwidth}
  \centering
 \begin{minipage}[t]{0.6\textwidth}
  \centering
  \subfigure[]{
	\includegraphics[width=.45\linewidth]{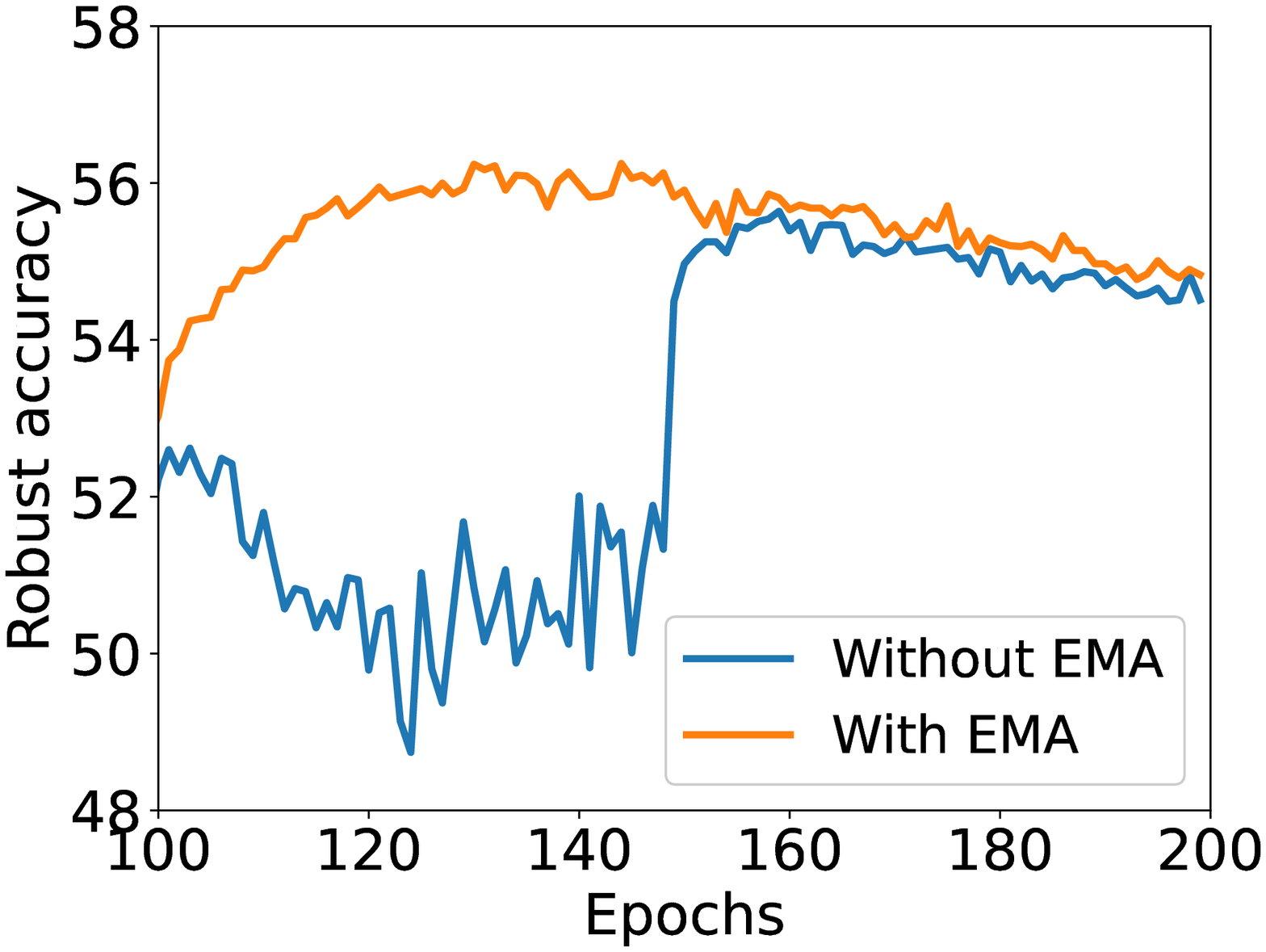}
	\label{fig5_a}
	}
	\subfigure[]{
	\includegraphics[width=.45\linewidth]{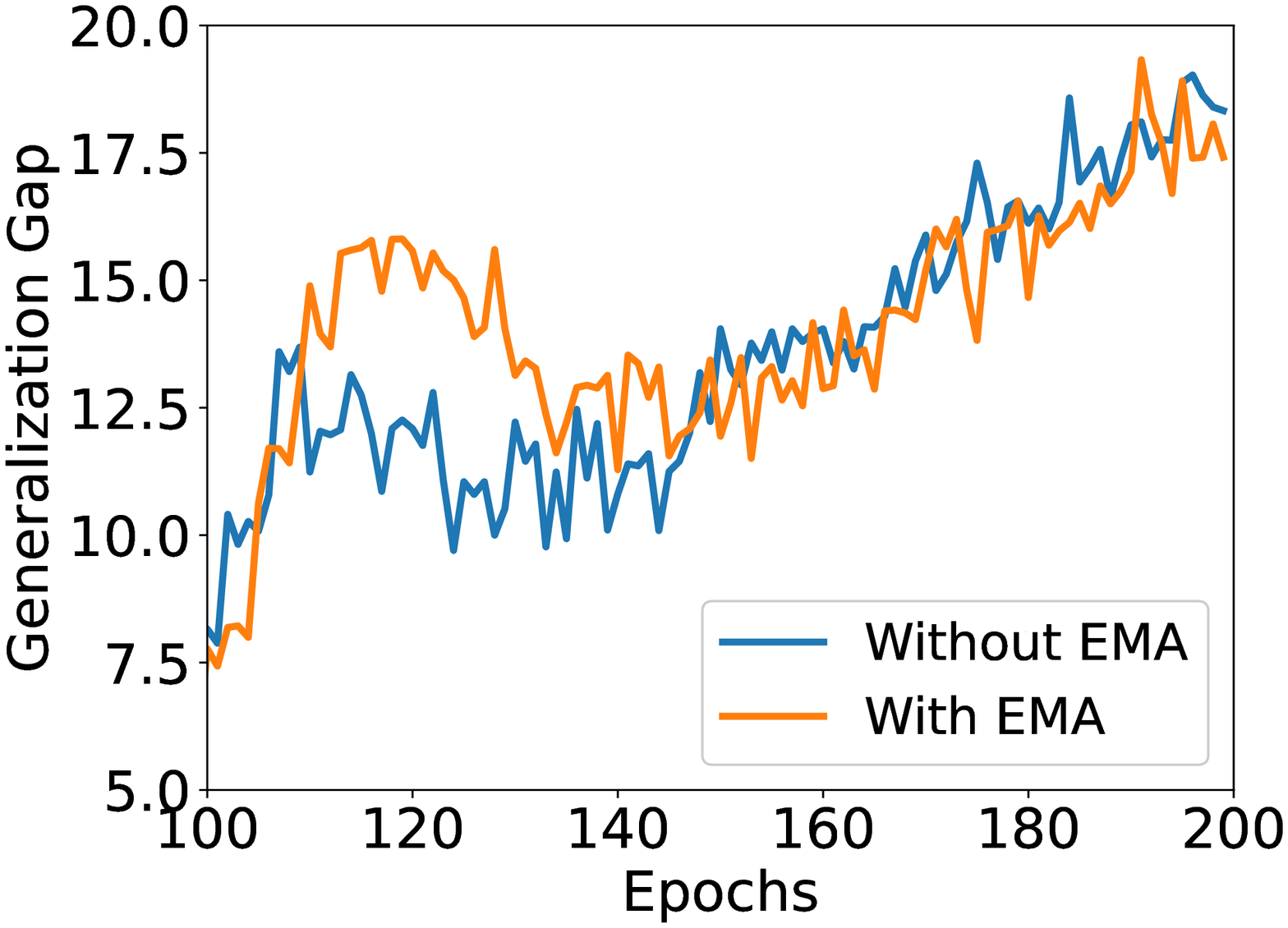}
	\label{fig5_b}
	}
  \vspace{-1em}
  \caption{\small \textbf{Impact of EMA. (a) Test robust; (b) Robust generalization gap}}
  \label{fig5}
  \vspace{1em}
 \end{minipage}

 \begin{minipage}[t]{0.6\textwidth}
  \centering
    \subfigure[]{
		\includegraphics[width=.45\linewidth]{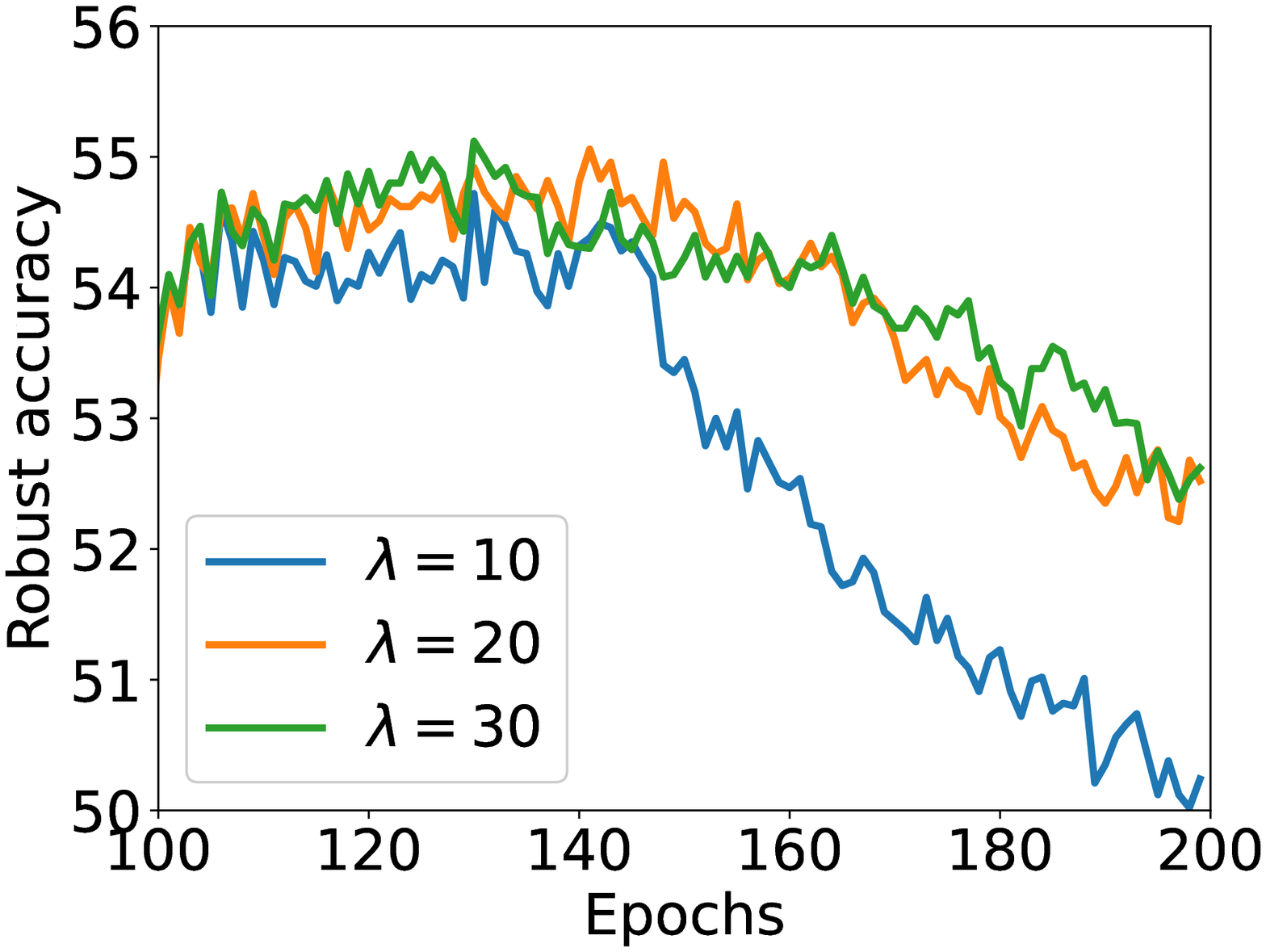}
		\label{fig6_a}
	}
    \subfigure[]{
	    \includegraphics[width=.45\linewidth]{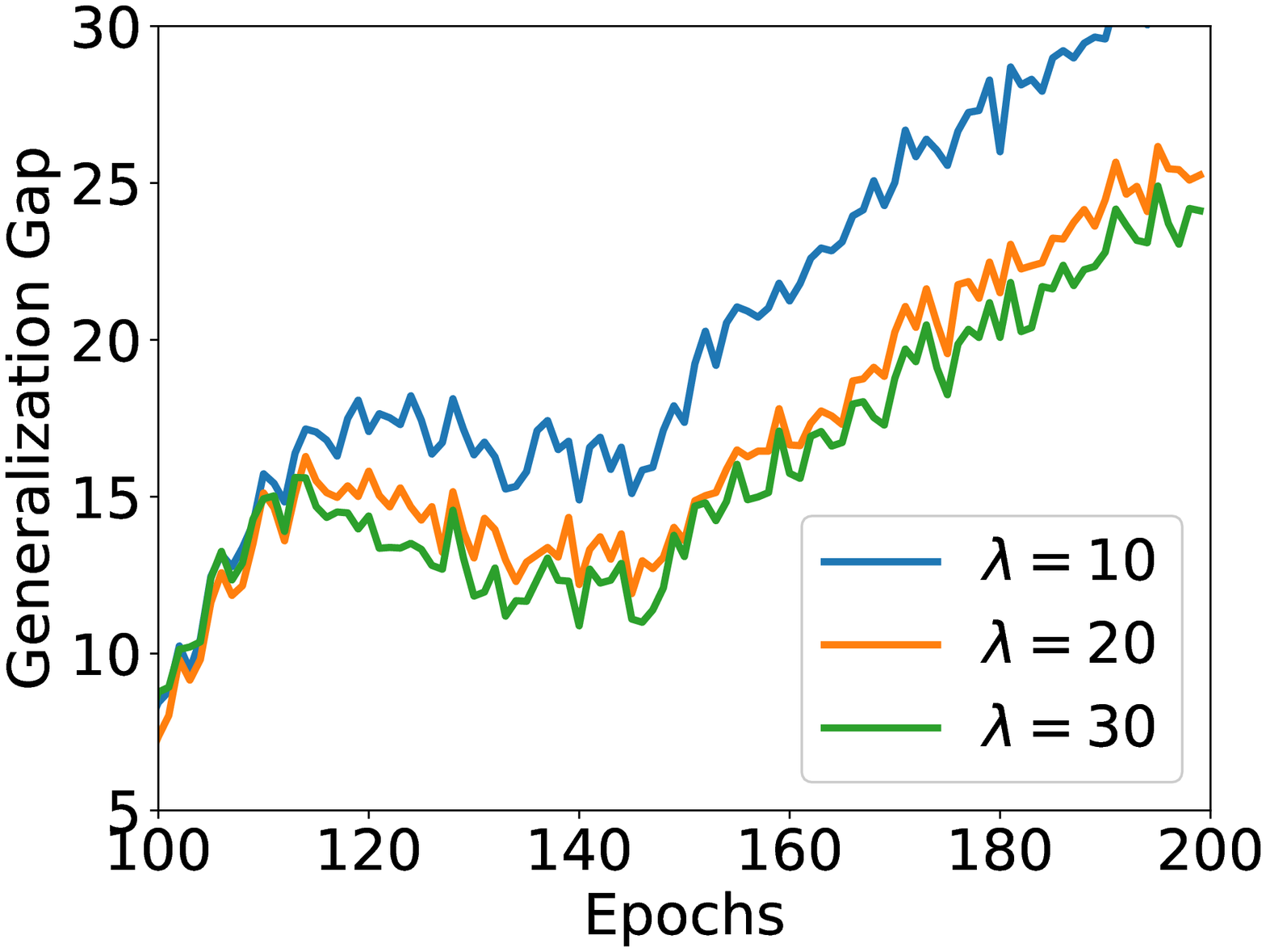}
	    \label{fig6_b}
	    }
	\vspace{-1em}
	\caption{\small\textbf{Impact of the consistency weight. (a) Test robust; (b) Robust generalization gap}}
	\label{fig6}
  \vspace{1em}
 \end{minipage}

 \begin{minipage}[t]{0.6\textwidth}
  \centering
  \subfigure[]{
	\includegraphics[width=.45\linewidth]{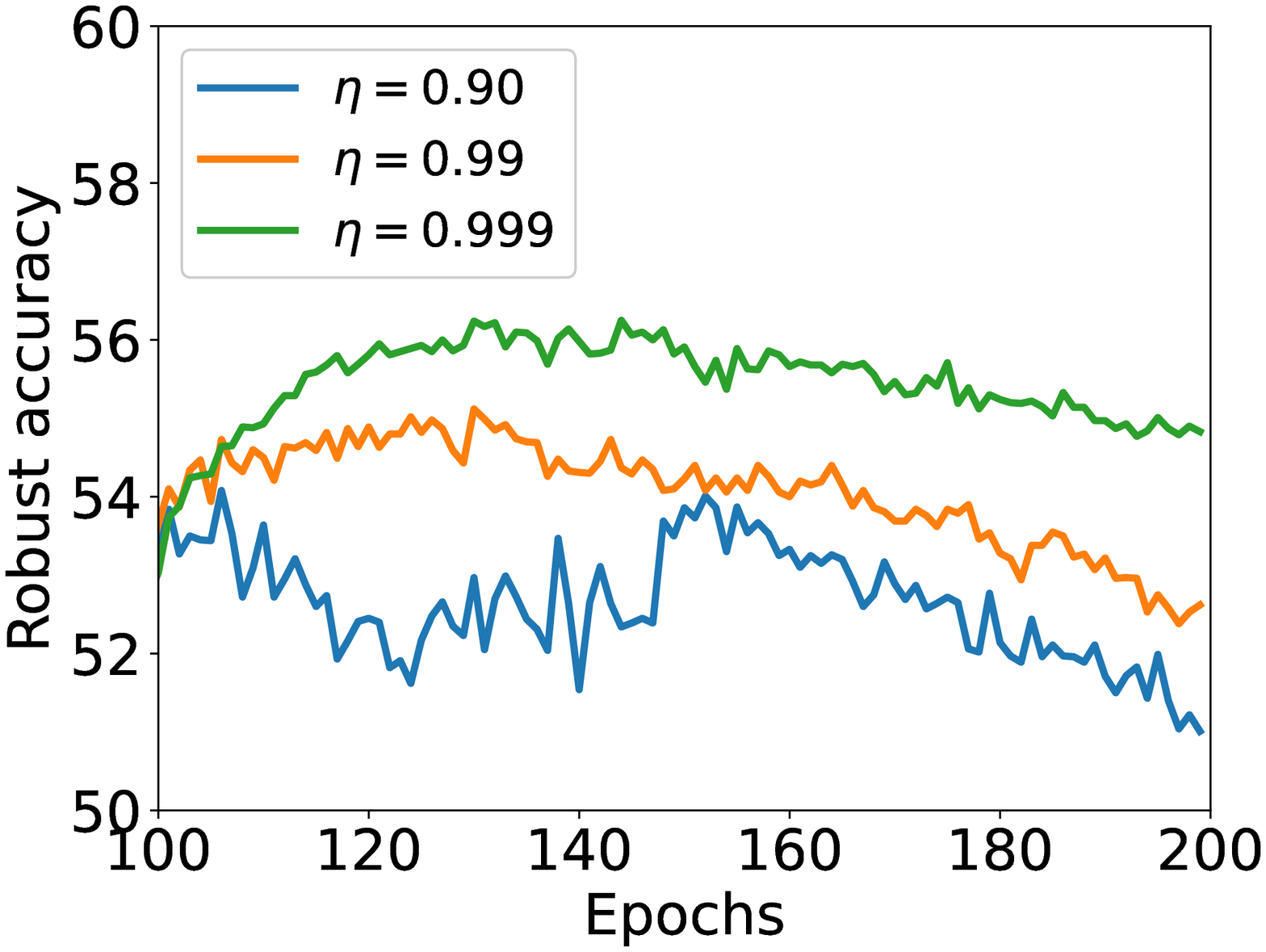}
	\label{fig7_a}
	}
	\subfigure[]{
	\includegraphics[width=.45\linewidth]{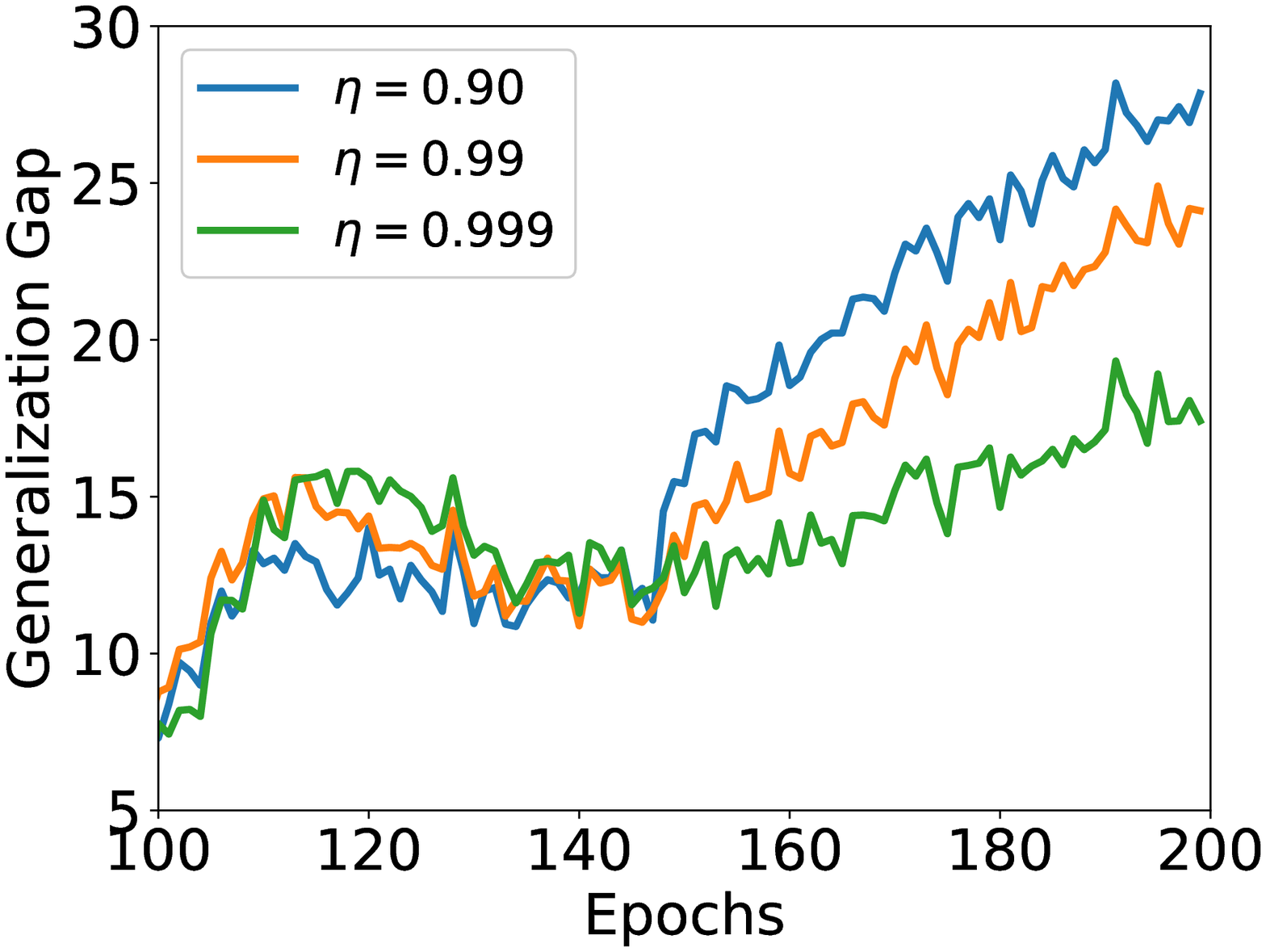}
	    \label{fig7_b}
	}
  \vspace{-1em}
  \caption{\small \textbf{Impact of the EMA decay. (a) Test robust; (b) Robust generalization gap}}
  \label{fig7}
    \vspace{-20pt}
 \end{minipage}
 \end{wrapfigure}

\noindent\textbf{Impact of EMA.}
We compare the robust accuracy and robust generalization gap without and with the EMA. The results are shown in Fig. \ref{fig5}. In the early stage after the first learning rate decay, compared with the student model without EMA, the robust accuracy of the teacher model on the test set is significantly improved. This is due to the fact that EMA can make the teacher model benefit from the memory of the student model. As the training progresses, the robust accuracy of the student model without EMA is gradually improved, and finally reaches similar performance as the teacher model. In terms of mitigating robust overfitting, there is almost no difference in the robust generalization gap between the teacher model with EMA and the student model without EMA, and both of them are much smaller than the vanilla PGD-AT. This indicates that the reduction of robust overfitting mainly depends on the consistency regularization rather than the EMA. The main contribution of EMA is to slightly improve the robust accuracy. 

\vspace{3pt}

\noindent\textbf{Impact of consistency weight and EMA decay.}
The hyper-parameters of the consistency weight $\lambda$ and EMA decay $\eta$ are critical in determining the effectiveness of our proposed method. Fig. \ref{fig6} shows the evaluation results with different values of these two hyper-parameters. We fix $\eta=0.99$ and vary $\lambda$ as 10, 20 and 30. We can see that a large $\lambda$ helps reduce the robust generalization gap, and increase the test robustness. In Fig. \ref{fig7}, we fix $\lambda=30$ and vary $\eta$ as 0.90, 0.99 and 0.999. We observe that the robust generalization gap decreases as $\eta$ increases, and $\eta=0.999$ gives the highest test robustness and the smallest robust generalization gap. This is because in the training process, after the first learning rate decay, the test robustness of the student model improves very slowly. So using a larger $\eta$ can result in a better teacher model with longer memory. Such teacher model can give a more accurate consistency loss. In sum, a larger consistency weight $\lambda$ and EMA decay $\eta$ can improve the robust accuracy and reduce robust overfitting.
\vspace{3pt}

\begin{table*}[t]
	\centering
	\caption{The impact of warm-up. ``WP-N'' and ``NWP-N'' denote the cases with and without warm-up at the N\textit{th} epoch respectively.}
	\label{table_warm_up}
	\resizebox{0.9\textwidth}{!}{
	\begin{tabular}{c|c|cccc|c}
			\Xhline{1pt}
			 \multirow{2}{*}{Training Strategy}      & Natural 		   	& \multicolumn{4}{c|}{Robust Accuracy}   & Generalization \\ \cline{3-6}
			                                        & Accuracy           & PGD-10 & PGD-100 & C\&W-100 & AA       & Gap\\\Xhline{1pt}
			 
			  WP-1 	  & 79.46   & 54.43  &54.12&51.97& 49.63 &\textbf{13.23} \\
			  WP-100     & \textbf{83.56}   & \textbf{56.25}  &\textbf{55.71}&\textbf{52.30}& \textbf{50.75} & 17.41 \\
			  NWP-100    & 81.75   & 55.94  &55.44&52.21& 50.55 &16.24 \\ \Xhline{1pt}
		          
	\end{tabular}%
	}
\end{table*}
\noindent\textbf{Impact of warm-up.}
We further investigate the effect of warm-up on mitigating robust overfitting. We fix $\eta=0.999$ and $\lambda=30$, and set different epochs with and without warmup. The results are shown in Table \ref{table_warm_up}. We observe that warm-up at the 100\textit{th} epoch gives better robustness and natural accuracy. We recommend adding warm-up in the late stage of training.

\section{Conclusions}
\label{sec6}
In this paper, we first hypothesize and verify that consistency regularization is critical in improving the model robustness and mitigating robust generalization gap for AT. Inspired by this observation, we propose a self-supervised learning method, which can be incorporated with existing AT approaches to resolve the robust overfitting issue. We introduce and update a teacher model as the Exponential Moving Average weights of the student model across multiple training steps. We further utilize the consistency loss to make the prediction distribution of the student model over AEs consistent with that of the teacher model over clean samples. Comprehensive experiments verify the effectiveness of our method in improving robustness and reducing robust generalization gap. 

As future work, we will study the theoretical connection between consistency regularization and robust overfitting, and explore other more effective consistency regularization.



%
%
\bibliographystyle{splncs04}
\bibliography{main}
\end{document}